\documentclass[11pt]{article}

\PassOptionsToPackage{numbers, compress}{natbib}
\usepackage[utf8]{inputenc}
\usepackage{amsmath}
\usepackage{amsfonts}
\usepackage{mathrsfs}
\usepackage{amssymb}
\usepackage{amsthm}

\usepackage{libertine}
\usepackage{wrapfig}
\usepackage{graphicx}
\usepackage{multirow}
\usepackage{epstopdf}
\usepackage{multicol}
\usepackage{subfigure}
\usepackage{booktabs}

\usepackage{algorithmic}
\usepackage{algorithm}
\usepackage[algo2e,linesnumbered]{algorithm2e}
\usepackage{stackengine}
\usepackage{graphicx}
\usepackage{multirow}
\usepackage{epstopdf}
\usepackage{multicol}
\usepackage{caption}
\usepackage{subfigure}
\usepackage{enumitem}
\usepackage{amsmath}
\usepackage{pifont}
\usepackage{natbib}
\usepackage{bbm}
\usepackage{makecell}
\usepackage{mathtools}
\usepackage[table]{xcolor}
\definecolor{mydarkred}{rgb}{0.6,0,0}
\definecolor{mydarkgreen}{rgb}{0,0.6,0}
\usepackage[colorlinks,
linkcolor=mydarkred,
citecolor=mydarkgreen]{hyperref}
\usepackage{url}
\usepackage{bm}
\usepackage{pifont}
\usepackage{stfloats}
\usepackage[T1]{fontenc}
\usepackage{arydshln}
\usepackage{colortbl}
\usepackage{balance}

\newcolumntype{L}[1]{>{\raggedright\let\newline\\\arraybackslash\hspace{0pt}}m{#1}}
\newcolumntype{Y}{>{\centering\arraybackslash}X}
\newcolumntype{s}{>{\hsize=.3\hsize}Y}
\newcolumntype{t}{>{\hsize=1.5\hsize}X}
\newcolumntype{u}{>{\hsize=0.8\hsize}Y}
\usepackage{makecell}
\usepackage{mathtools}
\usepackage[utf8]{inputenc}
\usepackage{stfloats}

\title{Instance Correction for Learning with \\Open-set Noisy Labels}

\author{
  Xiaobo Xia$^{1}$,
  Tongliang Liu$^{1}$,
  Bo Han$^{2}$,\\
  Mingming Gong$^3$,
  Jun Yu$^4$,
  Gang Niu$^5$,
  Masashi Sugiyama$^{5,6}$\\
  \small{$^1$The University of Sydney;}
  \small{$^2$Hong Kong Baptist University;}\\
  \small{$^3$The University of Melbourne;}
  \small{$^4$University of Science and Technology of China;}\\
  \small{$^5$RIKEN;}
  \small{$^6$The University of Tokyo}
}
\date{}
\begin{document}

\maketitle

\begin{abstract}
The problem of \textit{open-set noisy labels} denotes that part of training data have a different label space that does not contain the true class. Lots of approaches, e.g., loss correction and label correction, cannot handle such open-set noisy labels well, since they need training data and test data to share \textit{the same label space}, which does not hold for learning with open-set noisy labels. The state-of-the-art methods thus employ the \textit{sample selection} approach to handle open-set noisy labels, which tries to select clean data from noisy data for network parameters updates. The discarded data are seen to be mislabeled and do not participate in training. Such an approach is intuitive and reasonable at first glance. However, a natural question could be raised ``\textit{can such data only be discarded during training?}''. In this paper, we show that the answer is \textit{no}. Specifically, we discuss that the instances of discarded data could consist of some meaningful information for generalization. For this reason, we do not abandon such data, but use \textit{instance correction} to modify the instances of the discarded data, which makes the predictions for the discarded data consistent with given labels. Instance correction are performed by \textit{targeted adversarial attacks}. The corrected data are then exploited for training to help generalization. In addition to the analytical results, a series of empirical evidences are provided to justify our claims.
\end{abstract}

\newpage

\section{Introduction}\label{sec:1}
Noisy labels are ubiquitous in real-world data, which always arise in mistakes of manual or automatic annotators \cite{li2017webvision,li2020dividemix,xiao2015learning,yang2021free,wu2020class2simi,northcutt2021confident}. Learning with noisy labels can impair the performance of models, especially \textit{over-parameterized deep networks} which have large learning capacities and strong memorization power \cite{cheng2017learning,yao2020searching,zheng2020error,zhang2018generalized,jiang2020beyond,pleiss2020identifying}. Therefore, it is  of great importance to achieve robust training against noisy labels \cite{han2020sigua,han2018co,li2021provably,wang2019symmetric,menon2020can}.

The types of noisy labels studied so far can be divided into two categories: \textit{closed-set} and \textit{open-set} noisy labels. Closed-set noisy labels occur when instances have noisy labels that are \textit{contained within} the known class label set in the training data \cite{wang2018iterative}. On the other hand, open-set noisy labels occur when instances have noisy labels that are \textit{not contained within} the known class label set in the training data \cite{wang2018iterative}. A large body of work proposed various methods for learning with closed-set noisy labels, such as loss correction \cite{liu2019peer,patrini2017making,liu2016classification,xia2019revision,xia2020parts,yao2020dual,chen2019understanding,hendrycks2018using}, label correction \cite{tanaka2018joint,zheng2020error,zhang2021learning}, and sample selection \cite{han2018co,jiang2018mentornet,yu2019does,malach2017decoupling,wang2019co}. In contrast, learning with open-set noisy labels is less explored, which is \textit{our focus} in this paper. 

It is challenging to handle the open-set noisy label problem. Prior effects on learning with open-set noisy labels concentrated on the \textit{sample selection} approach \cite{wang2018iterative,yu2019does}. The reason for this is straightforward. That is to say, for open-set noisy labels, both loss correction and label correction are \textit{unreachable} since the true class label is unknowable for some training data \cite{wang2018iterative}. Consequently, prior effects have exploited sample selection to filter out mislabeled examples, and have used selected ``clean'' examples for robust training. The filtered examples are \textit{regarded as useless} and \textit{discarded} directly during the training procedure. 

Such a way of combating open-set noisy labels is \textit{intuitive but arguably suboptimal}, as the discarded data may contain meaningful information for generalization. For instance, let us consider the annotation process on the crowdsourcing platform \cite{yan2014learning,yu2018learning}, where the label space has been defined by experts. In this annotation process, the label ``dog" is \textit{within} the defined label space, but the label ``wolf'' is \textit{outside} the defined label space. Given an image, the annotators need to select one class label from the defined label space and assign it to this image. When there is an image of the wolf needed to be annotated, yet the label ``wolf'' is not within the label space, we have to select the label ``dog'' and assign it to the image of the wolf. The reason for this is that the \textit{semantic part label} \cite{xia2020parts} of the wolf image can be seen as ``dog'', since the wolf and dog have many similar features, e.g., ears, eyes, and legs. Such features are termed \textit{robust features}, which are \textit{comprehensible} to we humans \cite{ilyas2019adversarial}. Although the wolf image is mislabeled as ``dog'', robust features still are meaningful information for generalization. However, prior sample selection methods \cite{yu2019does,wang2018iterative,wang2019co} directly discard them. 

\begin{figure}[!t] 
\centering
\includegraphics[height=5.5cm, width=13.5cm]{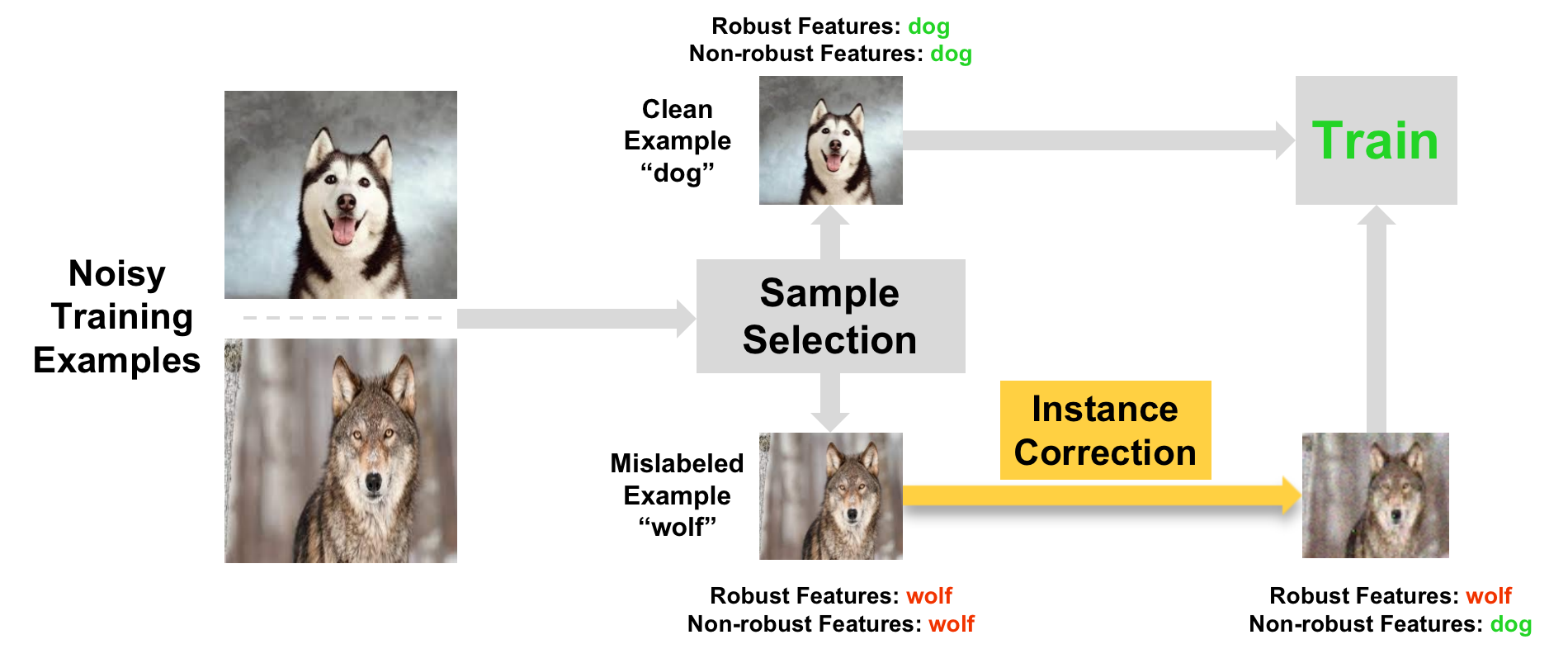}
\caption{\small{The illustration of the algorithm flow of the proposed method. When learning with open-set noisy labels, we first exploit the \textit{sample selection} approach to distinguish the clean example ``dog'' and mislabeled example ``wolf''. Then we employ targeted adversarial attacks to perform \textit{instance correction}, which reserves robust features of the wolf and changes non-robust features of the wolf to those of the dog. Finally, we utilize both the clean example and corrected example for training with the supervision of the label ``dog''.}}
\label{fig:main} 
\end{figure}

In this paper, to relieve the issue of ignoring meaningful information of discarded data in learning with open-set noisy labels, we perform \textit{instance correction} on discarded data to make use of them. Specifically, we employ \textit{targeted adversarial attacks} \cite{madry2017towards} which attempt to change the output classification of the input to a special target class. For the discarded data, the adversarial targeted attacks make their instances match given labels \textit{actively}. In this way, we can achieve \textit{instance correction} to combat open-set noisy labels. The illustration of our method is provided in Figure~\ref{fig:main}.

The proposed instance correction is motivated by the fact that the features of an instance can be divided into \textit{robust features and non-robust features} \cite{ilyas2019adversarial}. Compared with robust features which have been discussed, non-robust features are \textit{brittle and incomprehensible} to we humans. Training with robust or non-robust features can yield  great test accuracy \cite{ilyas2019adversarial}. Besides, non-robust features are easier to be changed with adversarial attacks than the robust ones \cite{ilyas2019adversarial}. For the instances with labels that are \textit{not contained within} the known class label set in the training data, they may have some robust features like those in clean data to help generalization as discussed. However, they may have different non-robust features compared with clean ones, which do not help generalization. Therefore, we use targeted adversarial attacks to change such non-robust features to match given labels.

Before delving into details, we summarize the main contributions of this paper in two folds. First, we identify the issues of prior approaches on learning with open-set noisy labels and propose instance correction to relieve the issue of ignoring meaningful information in discarded data. Second, we conduct a series of experiments to justify our claims well. The rest of this paper is organized as follows. In Section \ref{sec:2}, we introduce the background of learning with open-set noisy labels. In Section \ref{sec:3}, we present how to make use of discarded data to enhance deep networks step by step. Experimental results are provided in Section \ref{sec:4}. Finally, we conclude the paper in Section \ref{sec:5}.

\section{Learning with Open-set Noisy Labels}\label{sec:2}
\textbf{Notations.} Vectors and matrices are denoted by bold-faced letters. We use $\|\cdot\|_p$ as the $\ell_p$ norm of vectors or matrices. Let $[z]=\{1,2,\ldots,z\}$. 

\textbf{Preliminaries.} Consider a classification task, where there are $c$ classes in total. Let $\mathcal{X}$ and $\mathcal{Y}$ be the instance space and label space respectively, where $\mathcal{X}\in\mathbbm{R}^d$ with $d$ being the dimensionality, and $\mathcal{Y}=[c]$. In traditional supervised learning, the given training labels are clean. By employing such a clean dataset $S=\{(\mathbf{x}_i,y_i)\}_{i=1}^n$, where $n$ denotes the sample size or the number of training data, our goal is to learn a classifier that can assign labels precisely for given instances. However, in many real-world applications, the clean labels cannot be observed and the observed labels are noisy. Consider a subset of $S$ (denoted by $S'$) is corrupted and denote the corrupted version of the subset $S'$ by $S^\star$. Let $\mathcal{Y}'$ be the label space of $S^\star$ (with $\mathcal{Y}'\cap\mathcal{Y}=\varnothing$)---this means that the instances in $S^\star$ no longer have labels in $\mathcal{Y}$. Therefore, the whole noisy dataset can be denoted by $\tilde{S}=(S\backslash S')\cup S^\star$. The aim is changed to learn a robust classifier that could assign clean labels to test data by exploiting the noisy dataset $\tilde{S}$.

\begin{figure}[!h]
\centering
\includegraphics[height=2.75cm, width=14cm]{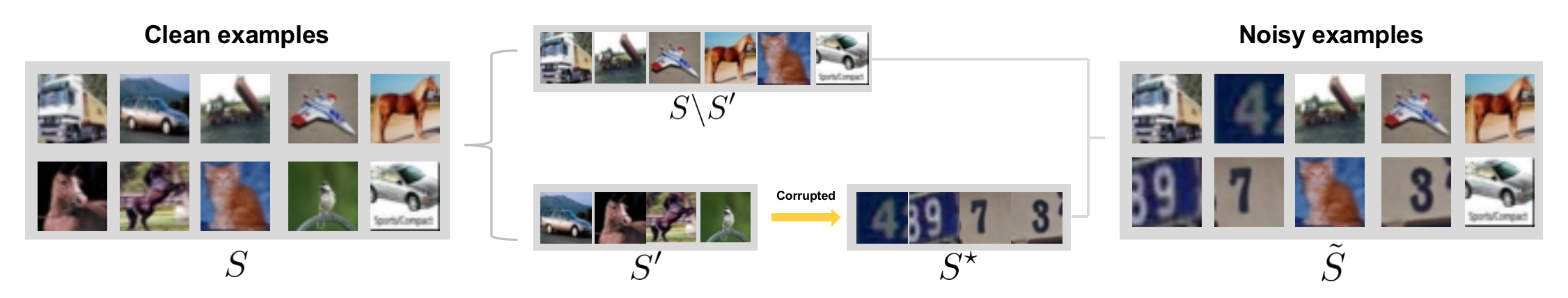}
\caption{The illustration of the used notations and problem setting.}
\label{fig:example} 
\end{figure}

Here, we provide an intuitive example for better understanding of the above notations (i.e., $S$, $S'$, $S^\star$, and $\tilde{S}$) and problem setting, which is shown in Figure \ref{fig:example}. We exploit the images of \textit{CIFAR-10} \cite{krizhevsky2009learning} and \textit{SVHN} \cite{netzer2011svhn}. More specifically, the clean examples come from \textit{CIFAR-10}, and the mislabeled examples (i.e., the examples in the set $S^\star$) come from \textit{SVHN}. In the generation process of open-set noisy labels, part of images of \textit{CIFAR-10} are corrupted by the images of \textit{SVHN}. Besides, the examples of \textit{CIFAR-10} and \textit{SVHN} do not have the same label space. For this reason, we only have a noisy sample set, i.e. $\tilde{S}$, and need to learn a robust classifier that can assign labels to test data.

\textbf{Limitations of loss/label correction approaches.} The methods of loss correction and label correction cannot handle noisy labels well \cite{wang2018iterative}. Here, we discuss the issues in more detail. For the loss correction approach, we need to model the flip processes from clean labels to noisy labels \cite{xia2019revision,yao2020dual,li2021provably,han2018masking,shu2020meta,liu2019peer}, which are \textit{within} the class label set of training data. When learning with open-set noisy labels, some instances do not have clean class labels which are within the class label set of training data. Therefore, when we tend to use the loss correction approach to handle open-set noisy labels, we will \textit{mistakenly} assign labels to mislabeled data. If this happens, the correction of the training losses will be incorrect, following bad classification performance.

For the label correction approach \cite{tanaka2018joint,zheng2020error,zhang2021learning}, we need to recalibrate the labels of mislabeled data based on the predictions of classifiers. As the classifier is trained only with the known class label set, it assigns labels to mislabeled data from this set. Apparently, such an assignment is incorrect, since the true labels of the mislabeled data are outside the known class label set. As a result, we still train the classifier on the dataset which contains incorrect labels. Hence, the obtained classifier cannot be robust under this circumstance.

\textbf{Shortcomings of the sample selection approach.} Prior effects exploited sample selection to handle open-set noisy labels \cite{wang2018iterative,yu2019does}, which only used the ``clean” examples (with relatively small losses) from each mini-batch for training. Such approaches inherit the \textit{memorization effects} of deep networks \cite{arpit2017closer}, which show that they would first memorize training data with clean labels and then those with noisy labels under the assumption that clean labels are of the majority in a noisy class. We use the self-teach model \cite{jiang2018mentornet,yao2020searching} as a representative example. The procedure is shown in Algorithm \ref{alg:gen}. Let $f$ be the classifier with learnable parameters. At the $t$-th iteration, when a mini-batch $\bar{S}$ is formed (Step 5), a subset of small-loss examples $\bar{S}_f$ is selected from the mini-batch $\bar{S}$ (Step 6). The size of $\bar{S}_f$ is determined by $R(T)$, which always depends on the noise rate $\tau$ \cite{han2018co}. Note that in \cite{han2018co}, the function $R(T)$ is designed as $1-\min\{{T}/{T_k}*\tau,\tau\}$. Such a design can help us make better use of the memorization effects of deep networks for sample selection. More specifically,  deep networks will learn clean and easy pattern in the initial epochs \citep{arpit2017closer}. Therefore, they have the ability to filter out mislabeled examples using their loss values at the beginning of training. When the
number of epochs goes large, they will eventually overfit to noisy labels \cite{han2018co}. For this reason,  at the start of training, we set $R(T)$ to a large value. Then, we gradually
increase the drop rate, i.e., reduce the value of $R(T)$. 

The selected ``clean” examples are then used to update the network parameters in Step 7. Those ``mislabeled” examples are \textit{discarded} directly and do not participate in training. To the end, since we select less noisy data for parameters updates, the classifier $f$ will be more robust. Note that in Step 6, we select small-loss examples $\bar{S}_f$ from $\bar{S}$ for parameter updates, where the examples in $\bar{S}\backslash\bar{S}_f$ are seen to be useless and discarded from training. Nevertheless, as discussed, such data could consist of meaningful information for generalization. It thus is not a great choice to abandon them directly during training, which makes the sample selection approach achieve sub-optimal performance. We discuss how to address the issue using the proposed method in next section.

\begin{algorithm}[!t]
\caption{General procedure on using sample selection to combat open-set noisy labels.}
\begin{algorithmic}[1]
	    \STATE \textbf{Input}: initialized classifier $f$, fixed $\tau$, epoch $T_k$ and $T_{\max}$, iteration $t_{\max}$.
		\FOR{$T = 0, \dots, T_{\max}-1$}
		\STATE \textbf{Shuffle} training set $\tilde{S}$;
		\FOR{$t = 0, \dots, t_{\max}-1$}
		\STATE \textbf{Draw} a mini-batch $\bar{S}$ from $\tilde{S}$;
		\STATE \textbf{Select} $R(T)$ small-loss examples $\bar{S}_f$ from $\bar{S}$ based on classifier's predictions;
		\STATE \textbf{Update} classifier parameter only using $\bar{S}_f$;
		\ENDFOR
		\STATE \textbf{Update} $R(T)=1-\min\{\frac{T}{T_k}\tau,\tau\}$;
		\ENDFOR
		\STATE \textbf{Output}: trained classifier $f$.
	\end{algorithmic}
	\label{alg:gen}
\end{algorithm}

\section{Method}\label{sec:3}
In this section, we introduce the proposed method in more detail. Recall Algorithm \ref{alg:gen}, the set of discarded data in each mini-batch can be denoted by $\bar{S}\backslash\bar{S}_f$. We focus on how to make use of the training data in $\bar{S}\backslash\bar{S}_f$ to help generalization.

Recent work \cite{ilyas2019adversarial} showed that adversarial examples can be directly attributed to the presence of \textit{non-robust features}, which are derived from patterns in the data distribution and \textit{highly predictive}. Note that non-robustness does not mean that the model is brittle, but part of features of images are \textit{easy to be changed via adversarial attacks}. Also, adversarial attacks mainly change the non-robust features \cite{ilyas2019adversarial}. Although non-robust features may be incomprehensible, they still can be used to 
\textit{achieve great classification performance}. Therefore, for our method, it is reasonable to \textit{generalize well} with corrected examples which are obtained by target adversarial attacks. We therefore can exploit targeted adversarial attacks on discarded data during training.

Let $\ell:\mathbbm{R}^c\times\mathcal{Y}\rightarrow\mathbbm{R}_{+}$ be a surrogate loss function for $c$-class classification. The loss function can be used to measure the degree of certainty that $f$ classifies the input to specific classes. In this paper, we exploit the \textit{softmax cross entropy loss}. Given a discarded example $(\mathbf{x}_i,\tilde{y}_i)\in\bar{S}\backslash\bar{S}_f$, we construct an adversarial example from a benign example $\mathbf{x}_i$ by adding a perturbation vector $\delta_{\mathbf{x}_i}$ by solving the following optimization problem:
\begin{equation}
    \min_{\delta_{\mathbf{x}_i}}\ell(f(\mathbf{x}_i+\delta_{\mathbf{x}_i}),\tilde{y}_i)\ \text{subject to}\ \|\delta_{\mathbf{x}_i}\|_2\textless \rho \ \text{and} \ f(\mathbf{x}_i+\delta_{\mathbf{x}_i})=\tilde{y_i}, 
\end{equation}
where $\rho$ is the \textit{budget} of targeted adversarial attacks \cite{bai2021improving,wu2020adversarial,ma2018characterizing,wang2019convergence}. When the classifier $f$ outputs probabilities $p_f(\cdot|\mathbf{x}_i)$ associated with each class, the used adversarial loss is $\ell(f(\mathbf{x}_i+\delta_{\mathbf{x}_i}),\tilde{y}_i)=-p_f(\tilde{y}_i|\mathbf{x}_i+\delta_{\mathbf{x}_i})$, which essentially maximizes the probability of classification into the class $\tilde{y}_i$. Here, we denote the \textit{corrected instance} via targeted adversarial attacks by $\tilde{\mathbf{x}}_i$, i.e., $\tilde{\mathbf{x}}_i=\mathbf{x}_i+\delta_{\mathbf{x}_i}$. The set that includes the pairs of the corrected instances and given labels, i.e., $(\tilde{\mathbf{x}}_j,\tilde{y}_j)$, is denoted as $\bar{S}_p$. 

In this way, although the clean label of the instance $\mathbf{x}_i$ is not within the class label set of the training data, the clean label of the corrected instance $\tilde{\mathbf{x}}_i$ can be seen as the given label $\tilde{y}_i$ from the perspective of classifier outputs. As such corrected instances can be highly predictive and help generalization, after targeted adversarial attacks, both the corrected examples and reserved examples in $\bar{S}_f$ are employed for network parameter updates. In this work, we seek a balance between corrected examples and reserved examples during training. Our objective function is given as follows:
\begin{equation}\label{eq:objective}
    \min L=\lambda * \ell_{(\mathbf{x}_i,\tilde{y}_i)\in\bar{S}_f}(f(\mathbf{x}_i),\tilde{y}_i)+(1-\lambda) * \ell_{(\tilde{\mathbf{x}}_j,\tilde{y}_j)\in\bar{S}_p}(f(\tilde{\mathbf{x}}_j),\tilde{y}_j),
\end{equation}
where $\lambda\ge 0$ is a hyperparameter to balance the contributions of corrected examples and reserved examples for training. More discussions about this hyperparameter are provided in Section \ref{sec:4}.

\textbf{Overall procedure.} We summarize the overall procedure of the proposed method based on the self-teach model \cite{jiang2018mentornet,yao2020searching}. Note that the proposed method can also be added to the methods which exploit two networks for sample selection, e.g., Co-teaching \cite{han2018co,yu2019does}. We employ the self-teaching model as a representative example to be compared with Algorithm \ref{alg:gen} more intuitively. We summarize the overall procedure of the proposed method in Algorithm \ref{alg:ours}. Specifically, we first use the sample selection approach to obtain an initialized classifier (Step 3). Then we divide all training examples into a clean set $\bar{S}_f$ and a mislabeled set $S\backslash \bar{S}_f$ (Step 5). After this, we perform \textit{instance correction} on the instances in $S\backslash \bar{S}_f$ by using targeted adversarial attacks to make them match given labels and obtain the set $\bar{S}_p$ (Step 6). The new training example set which consists of both the corrected examples and selected examples is obtained for parameter updates (Step 8). Comparing Algorithm \ref{alg:gen} with Algorithm \ref{alg:ours} further, we can see that the proposed method are simply implemented based on the prior sample selection procedure. Note that the hyperparameter $T_c$ can be determined by using a noisy validation set. Although the noisy validation set cannot perform as good as the clean one, it does not introduce the assumption that some clean data are available, which is more realistic \citep{nguyen2020self}. 
\begin{algorithm}[!t]
\caption{Overll procedure of the proposed method.}
	\begin{algorithmic}[1]
	    \STATE \textbf{Input}: initialized classifier $f$, fixed $\tau$, epoch $T_{c}$ and $T_{\max}$.
		\FOR{$T = 0, \dots, T_{c}-1$}
		    \STATE \textbf{Use} the sample selection approach as did in Algorithm \ref{alg:gen};
		\ENDFOR
		
		 \STATE \textbf{Divide} all training examples into clean and mislabeled examples using predictions of $f$;
		 \STATE \textbf{Perform} instance correction for the mislabeled examples;
		\FOR{$T = T_c, \dots, T_{\max}-1$}
		    \STATE \textbf{Exploit} both clean and corrected examples for training as did in Eq.~(\ref{eq:objective});
		\ENDFOR
		\STATE \textbf{Output}: trained classifier $f$.
	\end{algorithmic}
	\label{alg:ours}
\end{algorithm}

\section{Experiments}\label{sec:4}
In this section, we justify our claims from two folds. First, we conduct experiments on the benchmark dataset with synthetic open-set noisy labels (Section \ref{sec:4.1}). Second, we conduct experiments on the real-world dataset which contains open-set noisy labels (Section \ref{sec:4.2}). 

\subsection{Experiments on the benchmark dataset}\label{sec:4.1}
\textbf{Dataset.} \textit{CIFAR-10} \cite{krizhevsky2009learning} is employed to verify the effectiveness of the proposed method, which is popularly used for evaluation of noisy labels in the literature \cite{wang2018iterative,xia2021robust,yu2019does,thulasidasan2019combating,ma2020normalized,ma2018dimensionality}. \textit{CIFAR-10} has 10 classes of color images including 50,000 training images and 10,000 test images. The size of color images is 32$\times$32.

\begin{figure}[!h] 
\centering
\includegraphics[width=1.00\textwidth]{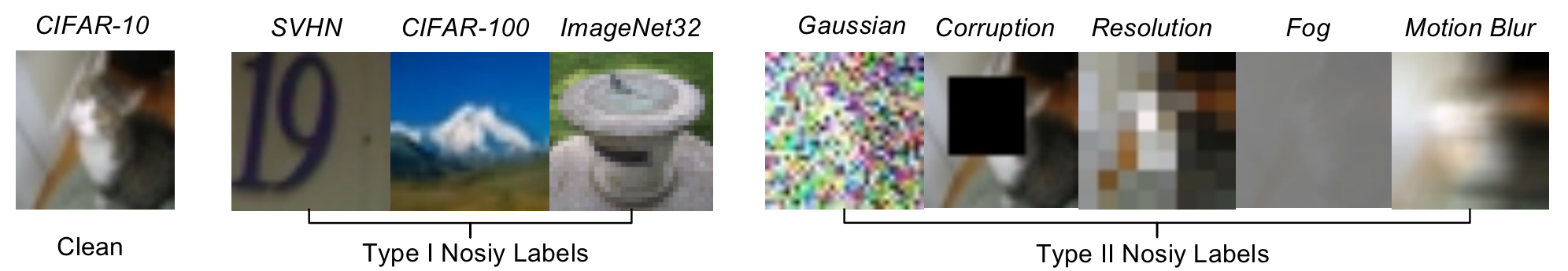}
\caption{Illustrations of examples of the noise injected to \textit{CIFAR-10}.}
\vspace{-10pt}
\label{fig:openset} 
\end{figure}

\textbf{Open-set noisy label generation.} Following \cite{wang2018iterative}, open-set noisy datasets are built by replacing some training images in \textit{CIFAR-10} by outside images, while keeping the labels and the number of images per class unchanged. We consider two types of noise, i.e., Type \uppercase\expandafter{\romannumeral1} and Type \uppercase\expandafter{\romannumeral2} noise. Type \uppercase\expandafter{\romannumeral1} noise includes images
from \textit{SVHN} \cite{netzer2011svhn}, \textit{CIFAR-100} \cite{krizhevsky2009learning}, and \textit{ImageNet32} (32$\times$32 ImageNet images) \cite{chrabaszcz2017downsampled}, and only those images whose
labels exclude the 10 classes in \textit{CIFAR-10} are considered. Type \uppercase\expandafter{\romannumeral2} noise includes images damaged by Gaussian random noise,  corruption, resolution distortion, fog distortion, and motion blur distortion. Some examples of the type \uppercase\expandafter{\romannumeral1} and type \uppercase\expandafter{\romannumeral2}  open-set noisy labels are given in Figure~\ref{fig:openset}. The overall noise rate $\tau$ is set to 20\%, 40\%, 60\%, and 80\%. We leave out 10\% of the noisy training data as a validation set for model selection.

\textbf{Baselines.} We compare the proposed method with the state-of-the-arts and the most related techniques for learning with open-set noisy labels: (1) Forward \cite{patrini2017making}, which estimates the noise transition matrix to combat noisy labels. (2) Joint \cite{tanaka2018joint}, which jointly optimizes the labels of examples and the network parameters for label correction. (3) SIGUA \cite{han2020sigua}, which exploits stochastic integrated gradient underweighted ascent to handle noisy labels. We use self-teach SIGUA in this paper. (4) MentorNet \cite{jiang2018mentornet}, which learns a curriculum to filter out noisy data. We used self-teach MentorNet in this paper. (5) S2E \cite{yao2020searching}, which exploits the AutoML technology to handle noisy labels. Note that the baseline (1) belongs to the loss correction approach. The baselines (2) belong to the label correction approach. Besides, the baselines (3), (4), and (5) belong the sample selection approach. 

Besides, we add a comparison called Mix, which directly involves discarded data into training without instance correction. Mix is based on the initialization of S2E. Our method, i.e., instance correction, is denoted as InsCorr in experiments. Since the proposed method is built on the sample selection approach, i.e., MentorNet (M) and S2E (S), we denote the proposed method as M-InsCorr and S-InsCorr respectively. Also, all methods Mix, M-InsCorr, and S-InsCorr are proposed in this paper, as prior works on sample selection do not make use of discarded data during training. We mark our methods with a symbol $\star$. Note that we do not compare our method with some state-of-the-art methods, e.g., SELF \cite{nguyen2020self} and DivideMix \cite{li2020dividemix}. It is because their proposed methods are aggregations of multiple techniques. The comparison with our method is thus not fair.

\textbf{Experimental setup.} For the fair comparison, we implement all methods with default parameters by PyTorch, and conduct all the experiments on NVIDIA Tesla V100 GPUs. A 7-layer CNN architecture is exploited following \cite{wang2018iterative,yu2019does}, which is standard test bed for weakly-supervised learning. The network architecture consists of 6 convolutional layers and 1 fully-connected layer. Batch normalization is applied in each convolutional layer before the ReLU activation, and a max-pooling layer is implemented every two convolutional layers. For all experiments, Adam optimizer (momentum=0.9) is with an initial learning rate of 0.001, and the batch size is set to 128. We run 200 epochs in total. As for targeted adversarial attacks, we employ targeted LinfPGD-Attacks \cite{madry2017towards} with the budget 8/255 and borrow the implementation of the Advertorch toolbox \cite{ding2019advertorch}\footnote{The official repository of Advertorch: https://github.com/borealisai/advertorch}.

For MentorNet and M-InsCorr, we set $R(T)=1-\min\{\frac{T}{T_k}\tau,\tau\}$. Here, we set $T_k$ to 10 as did in \cite{yu2019does}.  If the noise rate $\tau$ is not known in advanced, it can be inferred using validation sets \cite{liu2016classification,yu2018efficient}. For S2E and S-InsCorr, we set $R(T)$ as did in \cite{yao2020searching}, and exploit the official code \footnote{The official repository of S2E: https://github.com/AutoML-4Paradigm/S2E}.  Note that $R(T)$ only depends on the memorization effect of deep networks but not any specific datasets \cite{han2018co}. As for the performance measurement, we use the test accuracy, i.e., \textit{test accuracy = (\# of correct prediction) / (\# of testing)}. All experiments on \textit{CIFAR-10} are repeated five times. We report the mean and standard deviation of experimental results. Intuitively, if a method can achieve higher classification performance, it can better handle open-set noisy labels. 

\begin{table}[!t]
    \centering
    \small
    \begin{tabular}{cc|ccc|c|cc|cc}
         \Xhline{3\arrayrulewidth}
         \multicolumn{2}{c|}{Noise setting} & Forward & Joint & SIGUA & Mix$\star$ & MentorNet &M-InsCorr$\star$&S2E&S-InsCorr$\star$\\
         \hline
         \multirow{4}{*}{\textit{C}+\textit{S}}& 20\% &\makecell{74.26\\ $\pm$\scriptsize{0.49}} & \makecell{79.11\\ $\pm$\scriptsize{1.53}}&\makecell{70.88\\ $\pm$\scriptsize{2.33}}&\makecell{80.68\\ $\pm$\scriptsize{0.19}}&\makecell{82.77\\ $\pm$\scriptsize{0.30}}&\textbf{\makecell{82.81\\ $\pm$\scriptsize{0.60}}}&\makecell{82.32\\ $\pm$\scriptsize{0.80}}&\makecell{82.58\\ $\pm$\scriptsize{1.09}}\\
         \cdashline{2-10}[2pt/3pt]
         &40\%&\makecell{71.37\\ $\pm$\scriptsize{0.52}}&\makecell{74.71\\ $\pm$\scriptsize{1.78}}&\makecell{61.57\\ $\pm$\scriptsize{2.00}}&\makecell{75.71\\ $\pm$\scriptsize{0.56}}&\makecell{79.67\\ $\pm$\scriptsize{0.21}}&\textbf{\makecell{80.03\\ $\pm$\scriptsize{0.19}}}&\makecell{78.85\\ $\pm$\scriptsize{1.86}}&\makecell{79.16\\ $\pm$\scriptsize{1.35}}\\
         \cdashline{2-10}[2pt/3pt]
         &60\%&\makecell{53.83\\ $\pm$\scriptsize{2.37}}&\makecell{68.09\\ $\pm$\scriptsize{1.31}}&\makecell{54.71\\ $\pm$\scriptsize{2.04}}&\makecell{60.47\\ $\pm$\scriptsize{0.44}}&\makecell{72.68\\ $\pm$\scriptsize{3.95}}&\makecell{72.05\\ $\pm$\scriptsize{0.62}}&\textbf{\makecell{73.82\\ $\pm$\scriptsize{2.69}}} &\makecell{73.72\\ $\pm$\scriptsize{1.19}}\\
         \cdashline{2-10}[2pt/3pt]
         &80\%&\makecell{36.25\\ $\pm$\scriptsize{0.84}}&\makecell{39.06\\ $\pm$\scriptsize{3.12}}&\makecell{20.77\\ $\pm$\scriptsize{2.16}}&\makecell{59.61\\ $\pm$\scriptsize{0.64}}&\textbf{\makecell{64.90\\ $\pm$\scriptsize{2.19}}}&\makecell{62.35\\ $\pm$\scriptsize{0.32}}&\makecell{63.03\\ $\pm$\scriptsize{1.11}}&\makecell{62.36\\ $\pm$\scriptsize{1.55}}\\
         \hline
         \multirow{4}{*}{\textit{C}+\textit{C}}& 20\%& \makecell{76.72\\ $\pm$\scriptsize{0.93}}&\makecell{75.37\\ $\pm$\scriptsize{1.47}}&\makecell{74.27\\ $\pm$\scriptsize{0.16}}&\makecell{77.79\\ $\pm$\scriptsize{0.27}}&\textbf{\makecell{82.14\\ $\pm$\scriptsize{0.15}}}&\makecell{81.64\\ $\pm$\scriptsize{0.52}}&\makecell{80.34\\ $\pm$\scriptsize{0.79}}&\makecell{80.02\\ $\pm$\scriptsize{0.90}}\\
         \cdashline{2-10}[2pt/3pt]
         &40\%& \makecell{72.07\\ $\pm$\scriptsize{1.39}}&\makecell{67.30\\ $\pm$\scriptsize{0.39}}&\makecell{68.08\\ $\pm$\scriptsize{0.29}}&\makecell{68.79\\ $\pm$\scriptsize{0.26}}&\makecell{79.17\\ $\pm$\scriptsize{0.15}}&\textbf{\makecell{79.65\\ $\pm$\scriptsize{1.03}}}&\makecell{73.47\\ $\pm$\scriptsize{5.60}}&\makecell{74.52\\ $\pm$\scriptsize{2.73}}\\
         \cdashline{2-10}[2pt/3pt]
         &60\%& \makecell{54.95\\ $\pm$\scriptsize{1.06}}&\makecell{52.58\\ $\pm$\scriptsize{4.62}}&\makecell{53.45\\ $\pm$\scriptsize{1.46}}&\makecell{58.89\\ $\pm$\scriptsize{0.07}}&\makecell{74.58\\ $\pm$\scriptsize{0.31}}&\textbf{\makecell{74.86\\ $\pm$\scriptsize{0.28}}}&\makecell{60.77\\ $\pm$\scriptsize{7.46}}&\makecell{62.09\\ $\pm$\scriptsize{3.43}}\\
         \cdashline{2-10}[2pt/3pt]
         &80\%&\makecell{32.93\\ $\pm$\scriptsize{2.44}}&\makecell{42.95\\ $\pm$\scriptsize{1.29}}&\makecell{26.66\\ $\pm$\scriptsize{4.37}}&\makecell{40.41\\ $\pm$\scriptsize{0.89}}&\makecell{57.53\\ $\pm$\scriptsize{0.23}}&\textbf{\makecell{57.73\\ $\pm$\scriptsize{0.93}}}&\makecell{42.73\\ $\pm$\scriptsize{7.61}}&\makecell{46.24\\ $\pm$\scriptsize{2.65}}\\
         \hline
         \multirow{4}{*}{\textit{C}+\textit{I}}& 20\%&\makecell{77.12\\ $\pm$\scriptsize{0.38}}&\makecell{76.79\\ $\pm$\scriptsize{1.36}}&\makecell{72.16\\ $\pm$\scriptsize{0.92}}&\makecell{80.66\\ $\pm$\scriptsize{0.21}}&\makecell{82.15\\ $\pm$\scriptsize{0.77}}&\textbf{\makecell{82.47\\ $\pm$\scriptsize{0.06}}}&\makecell{80.25\\ $\pm$\scriptsize{1.02}}&\makecell{80.74\\ $\pm$\scriptsize{0.91}} \\
         \cdashline{2-10}[2pt/3pt]
         &40\%&\makecell{70.74\\ $\pm$\scriptsize{1.17}} &\makecell{65.76\\ $\pm$\scriptsize{2.64}}&\makecell{70.93\\ $\pm$\scriptsize{1.40}}&\makecell{73.47\\ $\pm$\scriptsize{1.01}}&\makecell{78.60\\ $\pm$\scriptsize{0.46}}&\textbf{\makecell{79.05\\ $\pm$\scriptsize{0.15}}}&\makecell{77.33\\ $\pm$\scriptsize{2.72}}&\makecell{77.87\\ $\pm$\scriptsize{2.24}}\\
         \cdashline{2-10}[2pt/3pt]
         &60\%& \makecell{60.30\\ $\pm$\scriptsize{2.56}}&\makecell{57.62\\ $\pm$\scriptsize{1.50}}&\makecell{59.54\\ $\pm$\scriptsize{1.65}}&\makecell{60.63\\ $\pm$\scriptsize{0.47}}&\textbf{\makecell{73.06\\ $\pm$\scriptsize{2.31}}}&\makecell{71.33\\ $\pm$\scriptsize{0.62}}&\makecell{64.09\\ $\pm$\scriptsize{2.88}}&\makecell{64.01\\ $\pm$\scriptsize{2.60}}\\
         \cdashline{2-10}[2pt/3pt]
         &80\%&\makecell{35.60\\ $\pm$\scriptsize{1.49}}&\makecell{37.09\\ $\pm$\scriptsize{4.06}}&\makecell{20.58\\ $\pm$\scriptsize{0.96}}&\makecell{40.09\\ $\pm$\scriptsize{3.61}}&\textbf{\makecell{58.06\\ $\pm$\scriptsize{3.76}}}&\makecell{56.24\\ $\pm$\scriptsize{2.24}}&\makecell{46.95\\ $\pm$\scriptsize{2.62}}&\makecell{49.62\\ $\pm$\scriptsize{3.04}}\\
         \Xhline{3\arrayrulewidth}
    \end{tabular}
    \vspace{5pt}
    \caption{Mean and standard deviation of
     test accuracy (\%) of different methods for learning with Type \uppercase\expandafter{\romannumeral1} open-set noisy labels over last ten epochs. ``\textit{C+S}'' is the abbreviation of \textit{\underline{C}IFAR-10}+\textit{\underline{S}VHN}. ``\textit{C+C}'' is the abbreviation of \textit{\underline{C}IFAR-10}+\textit{\underline{C}IFAR-100}.
    ``\textit{C+S}'' is the abbreviation of \textit{\underline{C}IFAR-10}+\textit{\underline{I}mageNet32}. The methods with the mark $\star$ are our methods. The best results are in bold.}
    \vspace{-8pt}
    \label{tab:type1}
\end{table}

\begin{table}[!t]
    \centering
    \small
    \begin{tabular}{cc|ccc|c|cc|cc}
         \Xhline{3\arrayrulewidth}
         \multicolumn{2}{c|}{Noise setting} & Forward & Joint & SIGUA & Mix$\star$ & MentorNet &M-InsCorr$\star$&S2E&S-InsCorr$\star$\\
         \hline
         \multirow{4}{*}{\textit{C}+\textit{G}}& 20\% &\makecell{74.62\\ $\pm$\scriptsize{0.88}}&\makecell{79.00\\ $\pm$\scriptsize{0.84}}&\makecell{75.97\\ $\pm$\scriptsize{0.21}}&\makecell{80.33\\ $\pm$\scriptsize{0.65}}&\makecell{80.56\\ $\pm$\scriptsize{0.20}}&\textbf{\makecell{81.17\\ $\pm$\scriptsize{0.43}}}&\makecell{79.53\\ $\pm$\scriptsize{2.14}}&\makecell{79.80\\ $\pm$\scriptsize{1.35}}\\
         \cdashline{2-10}[2pt/3pt]
         &40\% & \makecell{71.45\\ $\pm$\scriptsize{1.17}}&\makecell{75.11\\ $\pm$\scriptsize{0.12}}&\makecell{60.88\\ $\pm$\scriptsize{0.15}}&\makecell{75.15\\ $\pm$\scriptsize{0.35}}&\makecell{73.35\\ $\pm$\scriptsize{0.38}}&\makecell{76.32\\ $\pm$\scriptsize{0.27}}&\makecell{76.06\\ $\pm$\scriptsize{1.52}}&\textbf{\makecell{76.33\\ $\pm$\scriptsize{0.32}}}\\
         \cdashline{2-10}[2pt/3pt]
         &60\% & \makecell{60.62\\ $\pm$\scriptsize{1.86}}&\makecell{70.30\\ $\pm$\scriptsize{0.42}}& \makecell{37.69\\ $\pm$\scriptsize{5.62}}&\makecell{59.84\\ $\pm$\scriptsize{1.10}}&\makecell{56.93\\ $\pm$\scriptsize{0.68}}&\makecell{64.64\\ $\pm$\scriptsize{0.71}}&\makecell{71.80\\ $\pm$\scriptsize{1.46}}&\textbf{\makecell{72.25\\ $\pm$\scriptsize{1.83}}}\\
         \cdashline{2-10}[2pt/3pt]
         &80\%&\makecell{33.06\\ $\pm$\scriptsize{1.78}}&\makecell{53.17\\ $\pm$\scriptsize{3.94}}&\makecell{35.66\\ $\pm$\scriptsize{1.92}}&\makecell{53.98\\ $\pm$\scriptsize{1.77}}&\makecell{27.86\\ $\pm$\scriptsize{1.30}}&\makecell{48.44\\ $\pm$\scriptsize{3.05}}&\makecell{62.74\\ $\pm$\scriptsize{2.83}}&\textbf{\makecell{63.50\\ $\pm$\scriptsize{2.11}}}\\
         \hline
         \multirow{4}{*}{\textit{C}+\textit{O}}& 20\%& \makecell{75.79\\ $\pm$\scriptsize{1.05}}&\makecell{78.93\\ $\pm$\scriptsize{1.36}}&\makecell{66.33\\ $\pm$\scriptsize{0.31}}&\makecell{82.41\\ $\pm$\scriptsize{0.49}}&\makecell{81.81\\ $\pm$\scriptsize{0.31}}&\textbf{\makecell{82.55\\ $\pm$\scriptsize{0.35}}}&\makecell{79.47\\ $\pm$\scriptsize{2.96}}&\makecell{79.65\\ $\pm$\scriptsize{2.36}}\\
         \cdashline{2-10}[2pt/3pt]
         &40\%& \makecell{75.44\\ $\pm$\scriptsize{0.93}}&\makecell{75.82\\ $\pm$\scriptsize{1.25}}&\makecell{60.08\\ $\pm$\scriptsize{0.83}}&\makecell{80.17\\ $\pm$\scriptsize{0.02}}&\makecell{75.16\\ $\pm$\scriptsize{0.37}}&\makecell{77.51\\ $\pm$\scriptsize{0.09}}&\makecell{79.21\\ $\pm$\scriptsize{2.41}}&\textbf{\makecell{79.44\\ $\pm$\scriptsize{2.63}}}\\
         \cdashline{2-10}[2pt/3pt]
         &60\%&\makecell{64.05\\ $\pm$\scriptsize{0.87}}&\makecell{74.36\\ $\pm$\scriptsize{0.63}}&\makecell{30.08\\ $\pm$\scriptsize{1.00}}&\makecell{77.84\\ $\pm$\scriptsize{0.27}}&\makecell{57.09\\ $\pm$\scriptsize{0.60}}&\makecell{72.00\\ $\pm$\scriptsize{0.68}}&\makecell{77.73\\ $\pm$\scriptsize{1.80}}&\textbf{\makecell{77.90\\ $\pm$\scriptsize{1.08}}} \\
         \cdashline{2-10}[2pt/3pt]
         &80\%&\makecell{47.92\\ $\pm$\scriptsize{2.37}}&\makecell{67.32\\ $\pm$\scriptsize{0.88}}&\makecell{28.75\\ $\pm$\scriptsize{1.06}}&\makecell{69.54\\ $\pm$\scriptsize{0.25}}&\makecell{29.75\\ $\pm$\scriptsize{1.64}}&\makecell{64.75\\ $\pm$\scriptsize{0.91}}&\makecell{70.26\\ $\pm$\scriptsize{1.34}}&\textbf{\makecell{71.05\\ $\pm$\scriptsize{0.30}}}\\
         \hline
         \multirow{4}{*}{\textit{C}+\textit{R}}& 20\%&\makecell{77.46\\ $\pm$\scriptsize{0.94}}&\makecell{82.22\\ $\pm$\scriptsize{0.50}}&\makecell{66.19\\ $\pm$\scriptsize{0.22}}&\makecell{82.48\\ $\pm$\scriptsize{0.49}}&\makecell{82.58\\ $\pm$\scriptsize{0.12}}&\textbf{\makecell{82.91\\ $\pm$\scriptsize{0.13}}}&\makecell{79.18\\ $\pm$\scriptsize{1.41}}&\makecell{79.77\\ $\pm$\scriptsize{0.69}} \\
         \cdashline{2-10}[2pt/3pt]
         &40\%&\makecell{74.09\\ $\pm$\scriptsize{1.88}} &\makecell{78.80\\ $\pm$\scriptsize{1.56}}&\makecell{57.88\\ $\pm$\scriptsize{0.58}}&\makecell{79.00\\ $\pm$\scriptsize{0.16}}&\makecell{74.43\\ $\pm$\scriptsize{0.25}}&\textbf{\makecell{79.07\\ $\pm$\scriptsize{0.12}}}&\makecell{78.75\\ $\pm$\scriptsize{2.65}}&\textbf{\makecell{79.07\\ $\pm$\scriptsize{2.39}}}\\
         \cdashline{2-10}[2pt/3pt]
         &60\%&\makecell{65.63\\ $\pm$\scriptsize{1.05}}&\makecell{75.39\\ $\pm$\scriptsize{2.31}}&\makecell{28.84\\ $\pm$\scriptsize{1.86}}&\makecell{76.04\\ $\pm$\scriptsize{0.60}}&\makecell{56.91\\ $\pm$\scriptsize{0.84}}&\makecell{72.51\\ $\pm$\scriptsize{0.79}}&\makecell{76.65\\ $\pm$\scriptsize{2.24}}&\textbf{\makecell{77.25\\ $\pm$\scriptsize{2.21}}}\\
         \cdashline{2-10}[2pt/3pt]
         &80\%&\makecell{41.65\\ $\pm$\scriptsize{3.93}}&\makecell{52.65\\ $\pm$\scriptsize{1.79}}&\makecell{20.06\\ $\pm$\scriptsize{3.43}}&\makecell{65.19\\ $\pm$\scriptsize{1.87}}&\makecell{29.62\\ $\pm$\scriptsize{0.77}}&\makecell{62.91\\ $\pm$\scriptsize{1.15}}&\makecell{70.73\\ $\pm$\scriptsize{2.84}}&\textbf{\makecell{71.31\\ $\pm$\scriptsize{2.12}}}\\
         \hline
         \multirow{4}{*}{\textit{C}+\textit{F}}& 20\%&\makecell{75.45\\ $\pm$\scriptsize{0.93}}&\makecell{81.18\\ $\pm$\scriptsize{0.44}}&\makecell{70.09\\ $\pm$\scriptsize{0.43}}&\makecell{83.48\\ $\pm$\scriptsize{0.24}}&\makecell{82.70\\ $\pm$\scriptsize{0.23}}&\makecell{82.81\\ $\pm$\scriptsize{0.92}}&\makecell{83.15\\ $\pm$\scriptsize{1.34}}&\textbf{\makecell{83.52\\ $\pm$\scriptsize{1.37}}} \\
         \cdashline{2-10}[2pt/3pt]
         &40\%&\makecell{75.10\\ $\pm$\scriptsize{1.52}} &\makecell{78.91\\ $\pm$\scriptsize{0.56}}&\makecell{62.39\\ $\pm$\scriptsize{0.70}}&\makecell{81.05\\ $\pm$\scriptsize{0.24}}&\makecell{81.84\\ $\pm$\scriptsize{0.54}}&\makecell{82.03\\ $\pm$\scriptsize{0.09}}&\makecell{82.83\\ $\pm$\scriptsize{0.23}}&\textbf{\makecell{82.87\\ $\pm$\scriptsize{0.35}}}\\
         \cdashline{2-10}[2pt/3pt]
         &60\%&\makecell{66.08\\ $\pm$\scriptsize{0.83}}&\makecell{73.37\\ $\pm$\scriptsize{0.82}}&\makecell{40.24\\ $\pm$\scriptsize{0.95}}&\makecell{78.18\\ $\pm$\scriptsize{0.14}}&\makecell{76.72\\ $\pm$\scriptsize{0.32}}&\makecell{77.23\\ $\pm$\scriptsize{0.52}}&\makecell{78.76\\ $\pm$\scriptsize{0.17}}&\textbf{\makecell{79.03\\ $\pm$\scriptsize{1.09}}}\\
         \cdashline{2-10}[2pt/3pt]
         &80\%&\makecell{53.22\\ $\pm$\scriptsize{1.17}}&\makecell{68.06\\ $\pm$\scriptsize{1.78}}&\makecell{38.62\\ $\pm$\scriptsize{0.73}}&\makecell{69.89\\ $\pm$\scriptsize{0.73}}&\makecell{60.25\\ $\pm$\scriptsize{0.31}}&\makecell{61.02\\ $\pm$\scriptsize{0.39}}&\makecell{70.55\\ $\pm$\scriptsize{0.54}}&\textbf{\makecell{71.19\\ $\pm$\scriptsize{0.67}}}\\
         \hline
         \multirow{4}{*}{\textit{C}+\textit{M}}& 20\%&\makecell{77.79\\ $\pm$\scriptsize{0.75}}&\makecell{81.36\\ $\pm$\scriptsize{0.32}}&\makecell{68.50\\ $\pm$\scriptsize{0.23}}&\makecell{81.45\\ $\pm$\scriptsize{0.21}}&\makecell{82.61\\ $\pm$\scriptsize{0.42}}&\textbf{\makecell{82.66\\ $\pm$\scriptsize{0.14}}}&\makecell{79.98\\ $\pm$\scriptsize{1.32}}&\makecell{80.16\\ $\pm$\scriptsize{1.11}} \\
         \cdashline{2-10}[2pt/3pt]
         &40\%& \makecell{75.13\\ $\pm$\scriptsize{0.68}}&\makecell{78.24\\ $\pm$\scriptsize{1.05}}&\makecell{61.05\\ $\pm$\scriptsize{0.90}}&\makecell{78.56\\ $\pm$\scriptsize{0.19}}&\makecell{74.71\\ $\pm$\scriptsize{0.41}}&\makecell{77.30\\ $\pm$\scriptsize{0.92}}&\makecell{78.58\\ $\pm$\scriptsize{2.09}}&\textbf{\makecell{79.30\\ $\pm$\scriptsize{1.77}}}\\
         \cdashline{2-10}[2pt/3pt]
         &60\%& \makecell{64.75\\ $\pm$\scriptsize{1.62}}&\makecell{75.36\\ $\pm$\scriptsize{0.61}}&\makecell{33.44\\ $\pm$\scriptsize{2.60}}&\makecell{75.13\\ $\pm$\scriptsize{0.24}}&\makecell{57.34\\ $\pm$\scriptsize{2.38}}&\makecell{70.06\\ $\pm$\scriptsize{0.49}}&\makecell{76.34\\ $\pm$\scriptsize{1.85}}&\textbf{\makecell{76.49\\ $\pm$\scriptsize{2.68}}}\\
         \cdashline{2-10}[2pt/3pt]
         &80\%&\makecell{60.40\\ $\pm$\scriptsize{1.08}}&\makecell{64.39\\ $\pm$\scriptsize{2.82}}&\makecell{28.77\\ $\pm$\scriptsize{1.06}}&\makecell{68.58\\ $\pm$\scriptsize{0.64}}&\makecell{28.65\\ $\pm$\scriptsize{1.62}}&\makecell{61.72\\ $\pm$\scriptsize{0.66}}&\makecell{69.36\\ $\pm$\scriptsize{1.25}}&\textbf{\makecell{71.02\\ $\pm$\scriptsize{0.63}}}\\
         \Xhline{3\arrayrulewidth}
    \end{tabular}
    \vspace{5pt}
    \caption{Mean and standard deviation of
     test accuracy (\%) of different methods for learning with Type \uppercase\expandafter{\romannumeral2} open-set noisy labels over last ten epochs. ``\textit{C+G}'' is the abbreviation of \textit{\underline{C}IFAR-10}+\textit{\underline{G}aussian}. ``\textit{C+O}'' is the abbreviation of \textit{\underline{C}IFAR-10}+\textit{C\underline{o}rruption}.
    ``\textit{C+R}'' is the abbreviation of \textit{\underline{C}IFAR-10}+\textit{\underline{R}esolution}. ``\textit{C+F}'' is the abbreviation of \textit{\underline{C}IFAR-10}+\textit{\underline{F}og}. ``\textit{C+M}'' is the abbreviation of \textit{\underline{C}IFAR-10}+\textit{\underline{M}otion Blur}. The methods with the mark $\star$ are our methods. The best results are in bold.}
    \vspace{-8pt}
    \label{tab:type2}
\end{table}
\subsubsection{Analyses of experimental results}
\textbf{Overall results.} The experimental results with Type \uppercase\expandafter{\romannumeral1} and Type \uppercase\expandafter{\romannumeral2} open-set noise are provided in Table \ref{tab:type1} and \ref{tab:type2} respectively. With Type \uppercase\expandafter{\romannumeral1} open-set noise, the sample selection approach (e.g., MentorNet) always outperforms other types of approaches. When the noise level is high, the advantage is very obvious. For example, when the noise rate is 80\%, MentorNet always achieves a lead of more than 10\% over Joint. In addition, MentorNet achieves a lead of more than 20\% over Forward. The experimental results show the superiority of the sample selection approach in combating open-set noisy labels. 

With Type \uppercase\expandafter{\romannumeral2} open-set noise, the sample selection approach S2E achieves the best classification performance in most cases. When the noise level is high, the performance of another sample selection approach MentorNet is not promising. The reason is that MentorNet exploits a fixed sample selection procedure, i.e., $R(T)$. Also, when the noise level is high, it easily chooses the mislabeled examples, since the mislabeled ones are more similar with original ones in Type \uppercase\expandafter{\romannumeral2} open-set noise. What's worse, once MentorNet makes the wrong choice during training, the errors will be accumulated, which seriously hurts generalization \cite{yu2019does}. By contrast, S2E exploits the AutoML technique to control the sample selection procedure, which better exploits the memorization effects of deep networks, following better classification performance than MentorNet. Although SIGUA also focuses on sample selection, it uses stochastic integrated gradient underweighted ascent to handle open-set noisy labels, which highly relies on suitable selection of hyperparameters. Therefore, SIGUA does not achieve competitive performance with other sample selection methods.

\textbf{Mix vs. loss/label correction.} As shown in Table \ref{tab:type1}, on \textit{C+S} and \textit{C+C}, Mix can outperform Forward and Joint in most cases. On \textit{C+I}, Mix always achieves better classification performance than Forward and Joint. Such experimental results show the vulnerability of the loss and label correction approaches to handle open-set noisy labels. Similarly, as shown in Table \ref{tab:type2}, Mix can obtain better results compared with Forward and Joint in most circumstances.

\textbf{InsCorr vs. sample selection.} We first analyze the results with Type \uppercase\expandafter{\romannumeral1} open-set noise. In the most cases, the proposed instance correction can bring positive effects for generalization, which show the effectiveness of the proposed method. In some cases, instance correction cannot bring better performance. The reason is that Type \uppercase\expandafter{\romannumeral1} open-set noise is generated too randomly or even groundlessly. Therefore, the mislabeled examples consist of too many distinct robust features, which make it difficult to correct them by the proposed method. For instance, the images of \textit{CIFAR-10} is mainly about the objects in our daily life, e.g., cats, dogs, and frogs. However, the images of \textit{SVHN} is about the house numbers in Google Street View imagery. The images in two datasets are much different, which causes that instance correction sometimes does not work well. 

We then analyze the results with Type \uppercase\expandafter{\romannumeral2} open-set noise, where the noise is arguably generated more reasonably. In such cases, instance correction can change the non-robust features of mislabeled examples to match given labels, which brings better classification performance. The results justify our claims well, i.e., the discard data may have meaningful information, and can be used for training. 

\vspace{-5pt}
\subsubsection{Ablation study}
\begin{figure*}[t]
    \centering
    \begin{minipage}[c]{0.05\columnwidth}~\end{minipage}%
    \begin{minipage}[c]{0.3\textwidth}\centering\small  \textit{C+S}  \end{minipage}%
    \begin{minipage}[c]{0.3\textwidth}\centering\small \textit{C+C}  \end{minipage}%
    \begin{minipage}[c]{0.3\textwidth}\centering\small \textit{C+I}  \end{minipage}\\
    \begin{minipage}[c]{0.05\columnwidth}\centering\small \rotatebox[origin=c]{90}{-- Type \uppercase\expandafter{\romannumeral1} --} \end{minipage}%
    \begin{minipage}[c]{0.9\textwidth}
        \includegraphics[width=0.33\textwidth]{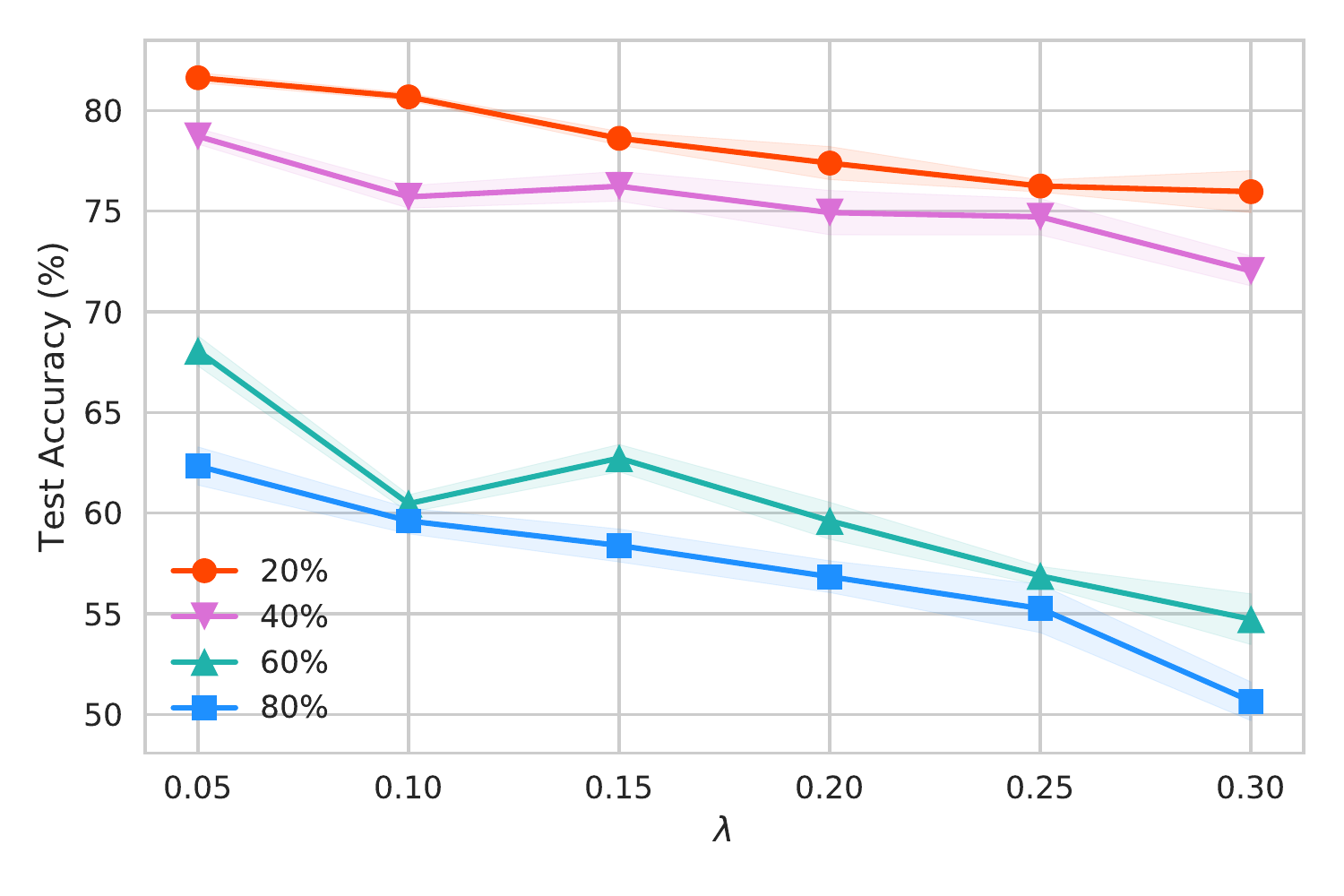}%
        \includegraphics[width=0.33\textwidth]{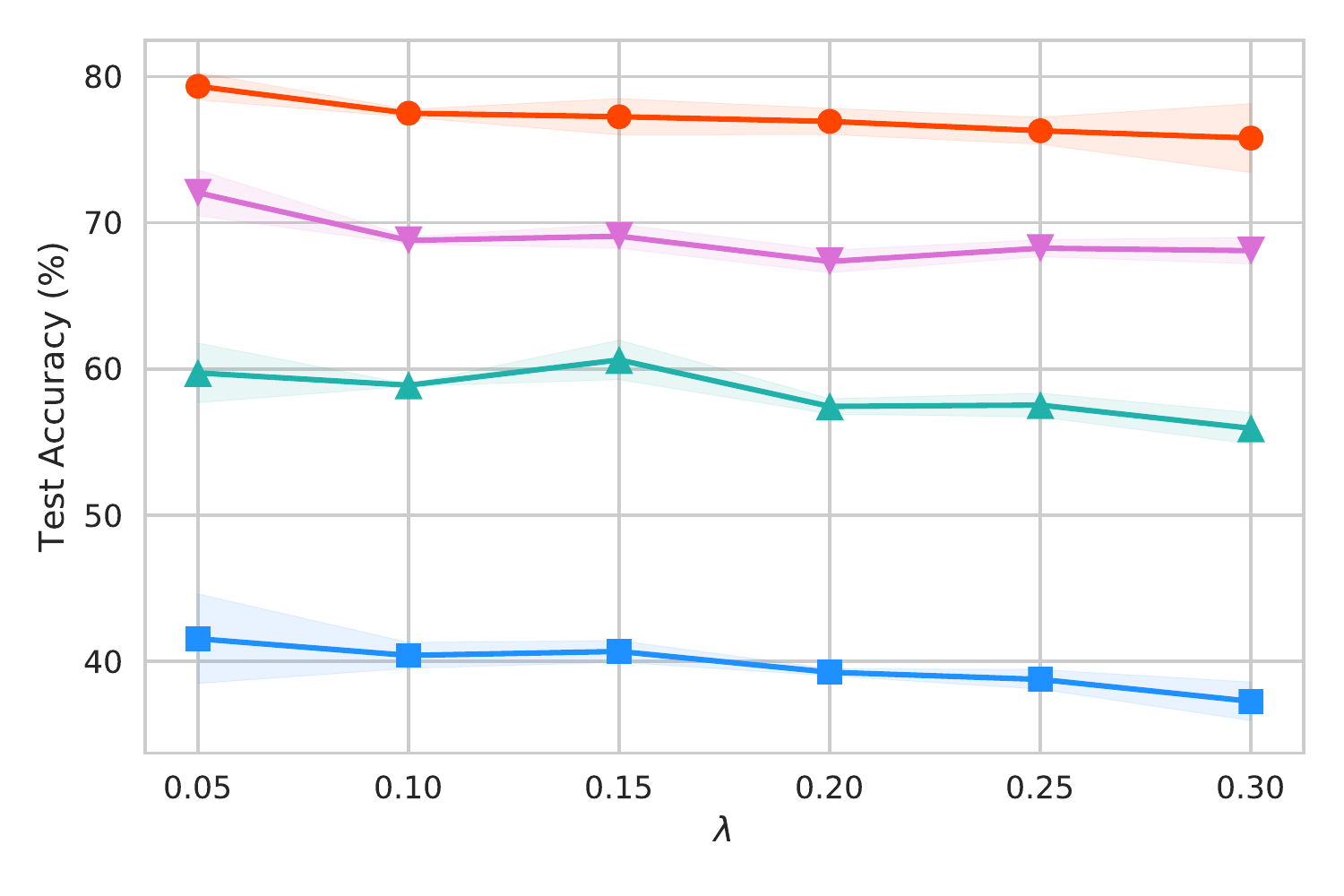}%
        \includegraphics[width=0.33\textwidth]{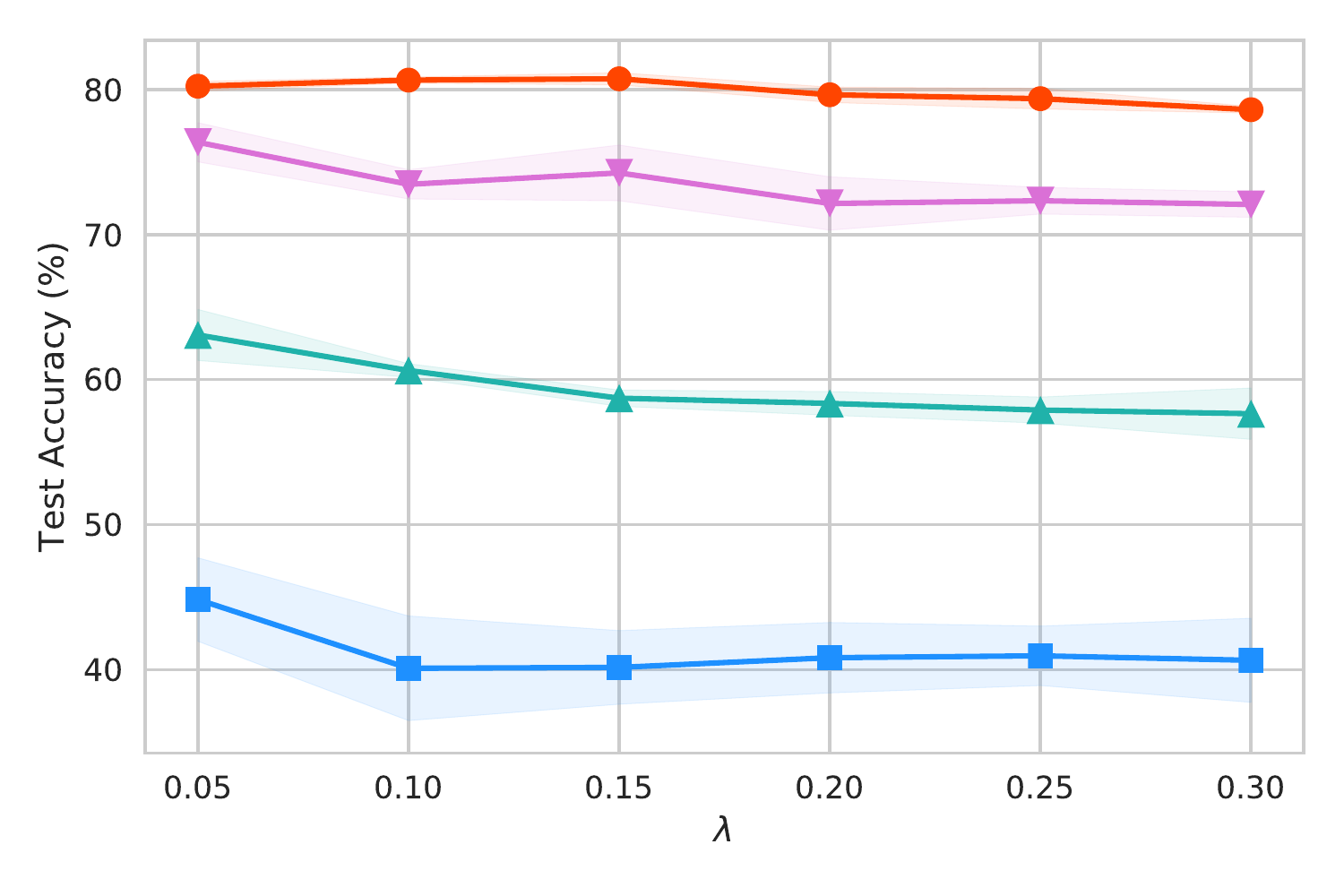}%
    \end{minipage}
    
    \begin{minipage}[c]{0.05\columnwidth}~\end{minipage}%
    \begin{minipage}[c]{0.3\textwidth}\centering\small  \textit{C+O}  \end{minipage}%
    \begin{minipage}[c]{0.3\textwidth}\centering\small \textit{C+R}  \end{minipage}%
    \begin{minipage}[c]{0.3\textwidth}\centering\small \textit{C+F}  \end{minipage}\\
    \begin{minipage}[c]{0.05\columnwidth}\centering\small \rotatebox[origin=c]{90}{-- Type \uppercase\expandafter{\romannumeral2} --} \end{minipage}%
    \begin{minipage}[c]{0.9\textwidth}
        \includegraphics[width=0.33\textwidth]{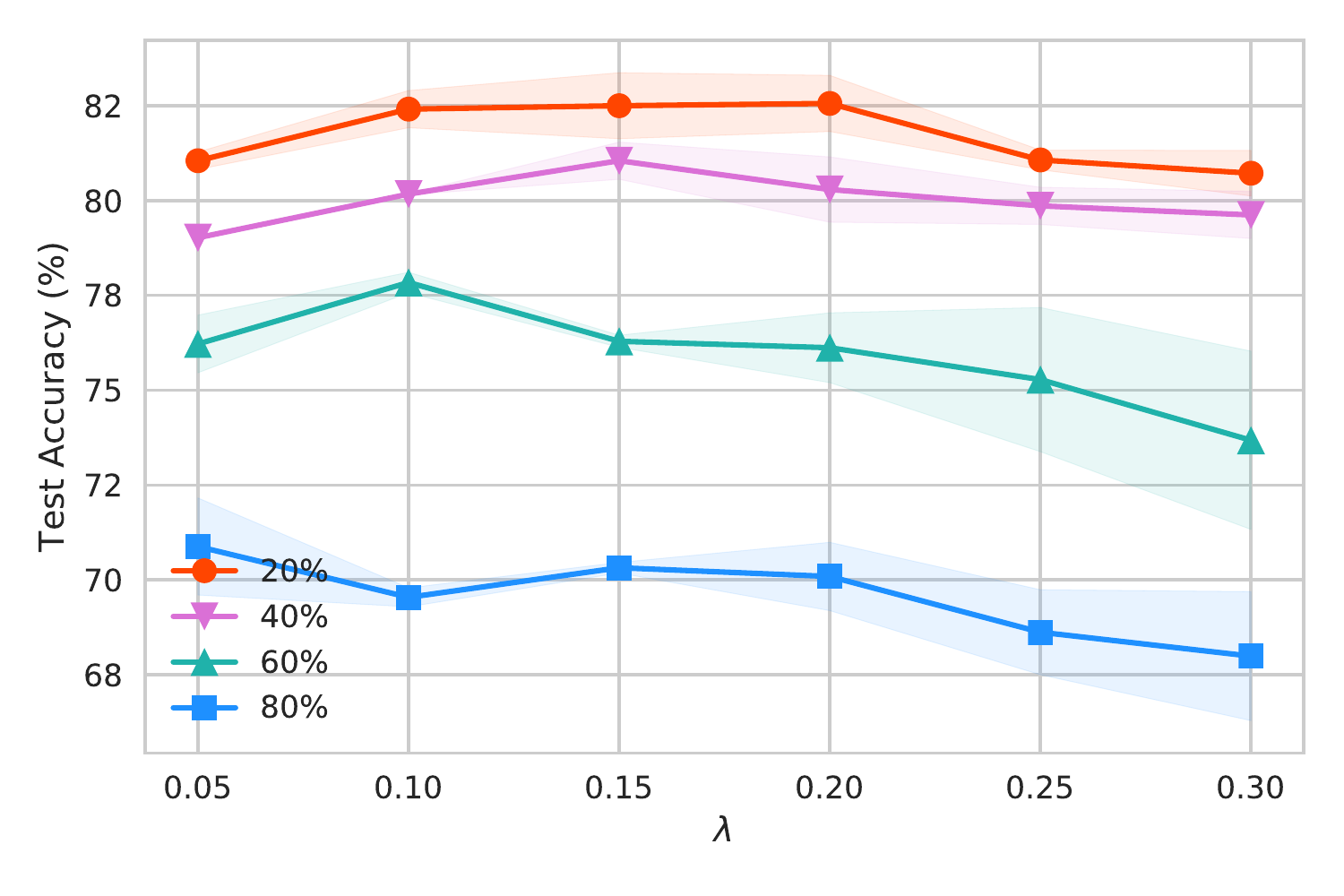}%
        \includegraphics[width=0.33\textwidth]{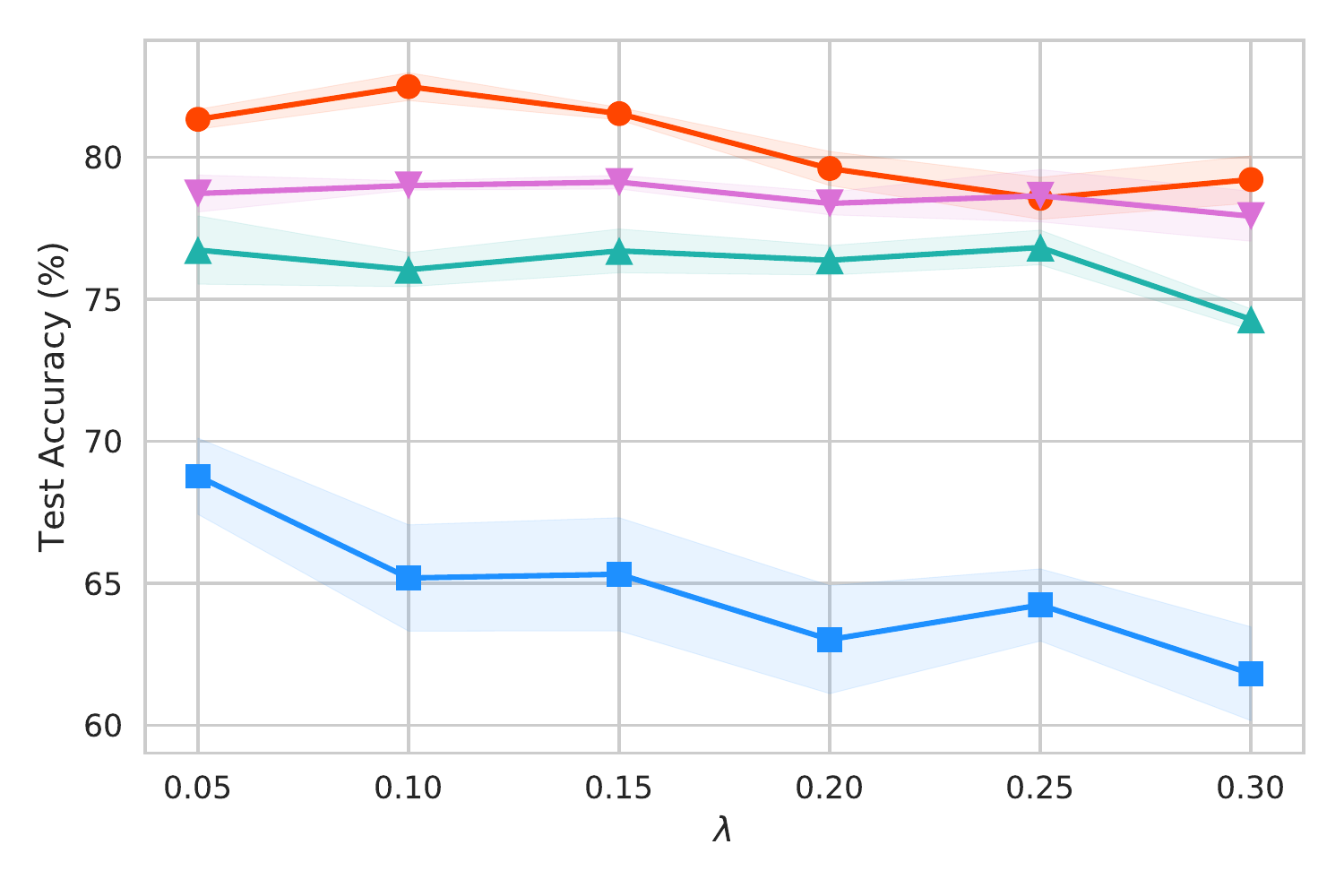}%
        \includegraphics[width=0.33\textwidth]{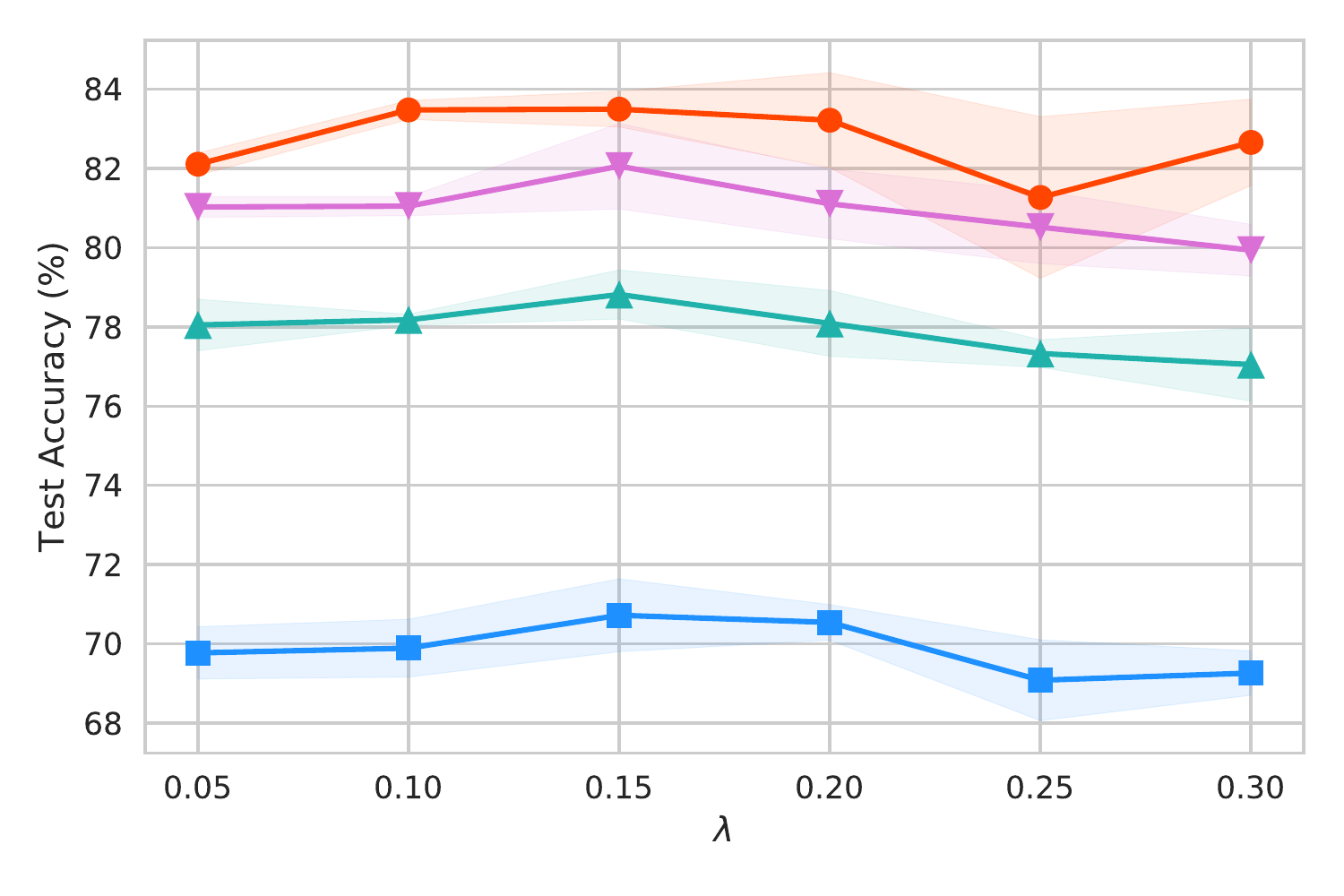}%
    \end{minipage}
    \caption{Illustrations of the test accuracy of the method Mix with different $\lambda$. The error bar for standard deviation in each figure has been shaded.}
    \label{fig:ab}
    \vspace{-5pt}%
\end{figure*}

We conduct an ablation study to analyze the influence of the hyperparameter $\lambda$ and show the impact of the discarded data. The experiments are conducted on the Type \uppercase\expandafter{\romannumeral1} noise datasets (i.e., \textit{C+S}, \textit{C+C}, and \textit{C+I}) and the Type \uppercase\expandafter{\romannumeral2} noise datasets (i.e., \textit{C+O}, \textit{C+R}, and \textit{C+F}). The experimental settings such as the network architecture and optimizer have been introduced before. The value of $\lambda$ is chosen from the range $\{0.05, 0.10, 0.15, 0.20, 0.25, 0.30\}$. The illustrations of results are provided in Figure~\ref{fig:ab}.

The discarded data have different influences in the experiments with Type \uppercase\expandafter{\romannumeral1} and Type \uppercase\expandafter{\romannumeral2} open-set noise. More specifically, with Type \uppercase\expandafter{\romannumeral1} open-set noise, we can see that the test accuracy decreases if we increase the the value of $\lambda$ in almost every case. Such results mean that the mislabeled examples in Type \uppercase\expandafter{\romannumeral1} cannot be directly involved into training, since they would hurt generalization. We need to first correct them and then can utilize them. As a comparison, with Type \uppercase\expandafter{\romannumeral2} open-set noise, the curve of test accuracy is not monotonically decreasing with the increase of $\lambda$. The results mean that the discarded data have positive effects to help generalization. Hence, we can make use of them appropriately. Note that such differences of results lie in the differences between the two noise generation mechanisms, which also show that the direct use of discarded data may be invalid.

\vspace{-5pt}
\subsection{Experiments on the real-world dataset}\label{sec:4.2}

\textbf{Experimental setup.} The \textit{WebVision} \citep{li2017webvision} dataset is employ in this paper. \textit{WebVision} contains 2.4 million images crawled from the websites using the 1,000 concepts in \textit{ImageNet ILSVRC12} \citep{deng2009imagenet}. Following the ``Mini'' setting in \citep{jiang2018mentornet,chen2019understanding,ma2020normalized}, we take the ﬁrst 50 classes of the Google resized image subset, and evaluate the trained networks on the same 50 classes of the \textit{ILSVRC12} validation set, which is exploited as a test set. Following prior works \citep{patrini2017making,xia2021robust}, we leave out 10\% training data as a validation set, which is for model selection. For \textit{WebVision}, we use an Inception-ResNet v2 \citep{szegedy2016inception} with batch size 128. The initial learning rate is set to $10^{-1}$. We set 100 epochs in total for all the experiments on the real-world dataset. 

\begin{table}[h]
	\centering
	
	{
		\begin{tabular}{ccc|c|cc|cc}
           	\Xhline{3\arrayrulewidth}
			 Forward & Joint & SIGUA & Mix$\star$&MentorNet&M-InsCorr$\star$&S2E &S-InsCorr$\star$\\
			 \hline
			 56.39 & 47.60 & 40.35 & 54.39 & 57.66 & \textbf{57.89} & 57.05 & 57.75\\
			 \Xhline{3\arrayrulewidth}
		\end{tabular}
	}
	\vspace{2pt}
	\caption{Top-1 validation accuracies (\%) on clean \textit{ILSVRC12} validation set of Inception-ResNet v2 models trained on \textit{WebVision}, under the ``Mini'' setting in \cite{jiang2018mentornet,chen2019understanding,ma2020normalized}. The best result is in bold.}
	\label{tab:webvision}
\end{table}
\textbf{Experimental results.} The experimental results are provided in Table \ref{tab:webvision}. Comparing M-InsCorr with MentorNet, we achieve an improvement of +0.23\%. Besides, comparing S-InsCorr with S2E, we achieve an improvement of +0.70\%. Note that directly involving discarded data can outperform the baselines such as Joint clearly, which means this type of approach is somewhat vulnerable in the practical problem. The results on the real-world dataset mean that the proposed method can handle classification tasks well in actual scenarios.
\section{Conclusion}\label{sec:5}
In this paper, we focus on the problem of learning with open-set noisy labels, where part of training data have a different label space that does not contain the true class. We first point out the weaknesses of the approaches such as loss correction, label correction, and sample selection for handling open-set noisy labels. Then we propose to use instance correction which is performed by targeted adversarial attacks to corrected the instances of discarded data to utilize them for training. A series of experimental results justify our claims well and verify the effectiveness of the proposed method. We believe that this paper opens up new possibilities in the topic of learning with open-set noisy labels. In the future, we will explore more ways to make use of discarded data and investigate more manners to perform instance correction to improve the robustness against open-set noisy labels. 

\section*{Acknowledgement}
TL was supported by Australian Research Council Project DE-190101473. BH was supported by the RGC Early Career Scheme No. 22200720 and NSFC Young Scientists Fund No. 62006202. JY was supported by USTC Research Funds of the Double First-Class Initiative (YD2350002001). GN was supported by JST AIP Acceleration Research Grant Number JPMJCR20U3, Japan. MS was supported by JST CREST Grant Number JPMJCR18A2, Japan.
\newpage
\bibliographystyle{plainnat}
\bibliography{bib}

\end{document}